\documentclass[journal]{IEEEtran}

\usepackage{cite}
\usepackage{natbib} % For bibliography
\usepackage{graphicx}
\usepackage{float} % H parametresi için
\usepackage{enumitem}
\usepackage{booktabs} % Daha güzel tablolar için
\usepackage{multirow}
\usepackage{colortbl}
\usepackage[utf8]{inputenc}
\usepackage{array}
\usepackage{soul, color}
\usepackage{multirow}
\usepackage{stfloats}
\usepackage{subcaption}
\usepackage{url}
\usepackage{geometry}
\begin{document}
\title{Deep Sequence Models for Predicting\\Average Shear Wave Velocity from\\Strong Motion Records}

\author{Baris~Yilmaz,
        Erdem~Akagündüz, Salih~Tileylioglu% <-this % stops a space

\thanks{Baris Yilmaz and Erdem Akagündüz are with the Dept. of Modeling and Simulation, Graduate School of Informatics, Middle East Technical University, Ankara, Türkiye. Salih Tileylioglu is with the Civil Engineering Department, Kadir Has University, İstanbul, Türkiye. }}

\maketitle

\begin{abstract}
This study explores the use of deep learning for predicting the time averaged shear wave velocity in the top 30 m of the subsurface ($V_{s30}$) at strong motion recording stations in Türkiye. $V_{s30}$ is a key parameter in site characterization and, as a result for seismic hazard assessment. However, it is often unavailable due to the lack of direct measurements and is therefore estimated using empirical correlations. Such correlations however are commonly inadequate in  capturing complex, site-specific variability and this motivates the need for data-driven approaches. In this study, we employ a hybrid deep learning model combining convolutional neural networks (CNNs) and long short-term memory (LSTM) networks to capture both spatial and temporal dependencies in strong motion records. Furthermore, we explore how using different parts of the signal influence our deep learning model. Our results suggest that the hybrid approach effectively learns complex, nonlinear relationships within seismic signals. We observed that an improved P-wave arrival time model increased the prediction accuracy of $V_{s30}$. We believe the study provides valuable insights into improving $V_{s30}$ predictions using a CNN-LSTM framework, demonstrating its potential for improving site characterization for seismic studies. Our codes are available via this repo: \url{https://github.com/brsylmz23/CNNLSTM_DeepEQ}
\end{abstract}

\begin{IEEEkeywords}
Shear wave velocity, strong ground motion records, deep learning prediction, hybrid model
\end{IEEEkeywords}

% \fi
\IEEEpeerreviewmaketitle

\section{Introduction}
Reducing seismic hazards and designing earthquake resistant structures are among the main objectives of earthquake engineering. Strong motion recordings play an important role in characterizing ground shaking hence shaping seismic design of structures. The recordings make up an important part of seismic hazard analysis as they provide direct measurements of ground shaking and help refine ground motion models and improve seismic design parameters. Ground motion models are fundamental parts of seismic hazard analysis and they rely on accurate characterization of local site conditions, as these conditions determine how earthquake waves are amplified or attenuated. The time averaged shear wave velocity over the upper 30 m of soil (\( V_{s30} \)) is a key parameter in assessing local site effects, as it influences how seismic waves propagate, the intensity of ground shaking, and the seismic demands on structures. 

$V_{s30}$ values which are typically determined by field measurements, are frequently unavailable at many recording sites around the world and existing proxies or correlations have shown limited accuracy and limited regional applicability. In this study, we explore the potential of utilizing a large dataset of earthquake recordings with deep learning techniques to predict $V_{s30}$ values at the earthquake recording stations in Türkiye.  

%Seismic activities such as earthquakes are measured through earthquake stations placed in certain regions. These measurements allow us to obtain information about earthquakes, and thanks to the analysis of this information, it becomes possible to design structures that are resistant to the destructiveness of earthquakes of similar magnitude, as well as precautions to be taken against earthquakes. The field of earthquake engineering, which carries out these analyses, is a critical branch of science that aims to ensure the safety of society and minimize the loss of life and property. One of the main goals of this field is to increase the durability of buildings against earthquakes, especially in countries located in earthquake zones. \\

%Türkiye is located in a critical earthquake zone due to its geographical structure. For example, events such as the 6.4 magnitude Hatay earthquake in 2023 and the 7.4 magnitude Marmara earthquake in 1999 reveal how sensitive Turkey is in terms of earthquake risk. This situation clearly shows the importance of increasing earthquake-related studies and precautions in Turkey. The aim of this study is to develop a hybrid modeling technique to estimate $V_{s30}$ values, which is a critical parameter of local soil types, in areas where strong motion stations of Turkish Disaster and Emergency Management (AFAD) are located in Turkey. The main motivation of this study is that earthquake waves vary depending on dynamic ground features and the need to model these differences accurately. 

\subsection{Problem Definition}
As stated in the previous section, this study aims to predict \( V_{s30} \) values at strong motion station sites using deep learning models trained on strong motion recordings. For this purpose, a dataset which includes recordings from various earthquake stations across Türkiye is constructed. These recordings are influenced by various factors from fault mechanisms to local site conditions, making them complex signals. Strong motion data are inherently non-stationary and generally consist of both low-and high-frequency components, which do not exhibit linear correlations with specific frequencies observable in these recordings. %What makes these records valuable is their nature as time series data, which encompass a range of components such as frequency, amplitude, and acceleration. Additionally, the peak ground acceleration (PGA), defined as the largest peak acceleration recorded on an accelerogram, represents one of the simplest and most fundamental strong-motion parameters \citep{douglas2001comprehensive}. 
Our main hypothesis is that the ability of deep learning models to learn complex, nonlinear relationships and hierarchical features inherent in such time-series data may provide an effective solution to this problem when applied to strong ground motion recordings.
%In addition to strong ground motion records, we also incorporate the arrival times of P-waves and S-waves, which we believe could provide high-level semantic information. This approach is considered to help extract meaningful information from complex time series data. \\

In our recent study \citep{yilmaz2024deep}, we investigated the potential of extracting features from strong ground motion records using convolutional layers. The goal was to evaluate whether convolutional neural networks (CNNs) could effectively capture relevant characteristics from these data. In the present study, we aim to extend this approach by framing it as a sequence problem. Specifically, we employ a hybrid system that combines CNNs with long short-term memory networks (LSTMs). While CNNs excel at extracting spatial features, LSTMs are adept at modeling complex sequential patterns in time-series data, making them effective for capturing the temporal dependencies in seismic signals. Additionally, we conducted experiments to analyze the impact of different signal segments on model performance, particularly focusing on the P-wave arrival time. These experiments aimed to determine the parts of the signal that contain the most informative features for analyzing the problem. By incorporating this insight, we optimized our hybrid approach to enhance feature extraction and improve the robustness of seismic event characterization.

We believe that employing a sequential model is essential because seismic signals as a type of time-series data, are inherently temporal and capture complex, time-dependent relationships. In the context of deep learning, sequence modeling has been extensively explored in the natural language processing (NLP) literature, where sequential data such as text or speech are used to understand {contextual relationships} over time. Notably, there are studies that attempt to apply the ability of NLP models to capture contextual relationships to time-series signals as well \citep{bian2024multi, zhou2023one}. Similarly, seismic signals exhibit sequential dependencies between complex patterns present within the strong motion signal. These temporal relationships may carry important information for accurately predicting parameters such as $V_{s30}$, which is the focus of this paper. LSTMs, with their ability to capture short or long-term dependencies, are particularly well-suited for this task, as they can effectively model the sequential nature of ground motion records. By combining CNNs, which excel at extracting complex patterns, with LSTMs, which specialize in modeling temporal dependencies, we hypothesize that this hybrid approach will effectively capture both the spatial and temporal characteristics of the data, thereby improving the accuracy of $V_{s30}$ predictions. To the best of our knowledge, this study represents the first attempt in the literature to address this problem using a hybrid CNN-LSTM approach.

The paper is structured as follows: Section 2 reviews the related literature and the dataset used, Section 3 details the experimental setup and model architecture, Section 4 presents the experimental results, Section 5 discusses the findings, and the final Section 6 concludes the paper with suggestions for future work.

\section{Literature Review}

Traditional methods for determining $V_{s30}$ include non invasive testing methods like the Refraction Microtremer (ReMi) \citep{louie2001faster}, Multi-Channel Analysis of Surface Waves (MASW) \citep{park1999multichannel}, as well as invasive approaches such as the 
Seismic Cross-Hole Test Seismic Testing \citep{stokoe1972situ}  and Seismic Downhole Testing \citep{robertson1986situ}. When geophysical measurements are unavailable, correlations such as penetration resistance \citep{Brandenberg2010}, surface geology, and topographic slope \citep{WaldandAllen2007} are used to estimate $V_{s30}$. There are also machine learning-based approaches that have investigated the problem. \cite{yaghmaei2011new} used two radial basis function neural networks to classify sites corresponding to NEHRP site classes (B, C, D, and E) based on horizontal-to-vertical spectral ratio (HVSR) curves. The study used data from 87 strong motion stations that recorded the 1999 Mw 7.6 Chi-Chi earthquake in Taiwan. The classification is carried out by calculating the HVSR curves from the recordings and comparing against reference HVSR curves developed for sites. HVSR curves can be used to estimate the fundamental site frequency \citep{nakamura2019nakamura}, which is a critical factor in understanding local site amplification and hence can be used in site classification. While using HVSR curves directly calculated from recordings makes the outlined procedure  efficient, the study has some limitations. Its use of ground motion data from only one earthquake, limited number of recordings and its reliance on HVSR curves potentially limit the applicability and generalization of the procedure to regions with different geological and seismic characteristics. \cite{yaghmaei2014site} provide a method to create site classification maps based on earthquake recordings. The study combines empirical and AI based site classification techniques to calculate site classification indices.  Kriging interpolation is used to create a spatial map of the site class indices for the region. The AI method used in the study is the the same as \cite{yaghmaei2011new}. The number of earthquakes used in the study were limited to two. Hence, the study shares similar shortcomings to \cite{yaghmaei2011new}.  

A study that directly predicts $V_{s30}$ values using Artificial Neural Networks (ANN) and Genetic Expression Programming (GEP) was carried out by \cite{gullu2013prediction}. In this work, \cite{gullu2013prediction}  applies the aforementioned methods to predict ($V_{s30}$) values by utilizing 84 strong ground motion records from 60 stations in California. The measured ($V_{s30}$)  values reported in the study ranges from 235 m/s to 902 m/s. The study also includes hand-crafted features such as earthquake magnitude, source-to-site distance, peak ground acceleration (PGA), and spectral acceleration as inputs. Overall, the ANN method achieved slightly better accuracy than the GEP method. It should be noted, however, that the study is limited by its relatively small dataset and its focus on a single region, which potentially restricts the generalization capability of its findings. The studies summarized so far do not incorporate architectures capable of extracting deep features. 

In our recent study \citep{yilmaz2024deep}, we demonstrated that convolutional neural networks effectively extract meaningful spatial features from strong ground motion records. One of the most important aspects of our study was the unique cross-validation approach we adopted to ensure effective regional testing. Unlike conventional methods, all recordings from the same station were kept exclusively in either the training or test set, never in both. This way, recordings from a station included in the training set were not present in the test set, and vice versa. This strategy was designed to evaluate features learned in one region against data from a different region, promoting a more robust assessment. Expanding on this, we now propose a hybrid CNN-LSTM system to model both spatial features and temporal dependencies in seismic signals, as no prior studies have attempted to predict $V_{s30}$ values from strong motion records using purely deep learning methods. Furthermore the dataset used in this study significantly surpasses those of previous studies investigating similar topics. 
\subsection{AFAD $V_{s30}$ Dataset}

In recent years, the number of earthquake datasets that can be used in artificial intelligence applications has increased significantly. However, the question of what data scale will be sufficient to generate high-level features in deep learning models is still a matter of significant debate \citep{ccauglar2024exploring}. For this reason, the Stanford Earthquake Dataset (STEAD) was introduced by \cite{mousavi2019stanford}, containing more than one million three-component seismograms of approximately 450 thousand earthquake events. Numerous significant studies have been conducted using the STEAD dataset. For example, \cite{Mousavi2020a} utilized the dataset for P-phase arrival time determination and epicenter location estimation, while \cite{Mousavi2020b} employed it to predict the P and S wave arrival times. Similarly, \cite{Ristea2022} used the STEAD dataset for earthquake location and magnitude estimation problems. It should be noted however, most of the signals within the STEAD dataset are not strong-motion records, and the dataset lacks information $V_{s30}$ values of the strong motion stations’ sites. Consequently, there is a need for a specialized dataset that addresses this gap to investigate problems such as one in the scope of this research.\\

\begin{figure}[t]
    \centering 
    \includegraphics[width=0.5\textwidth]{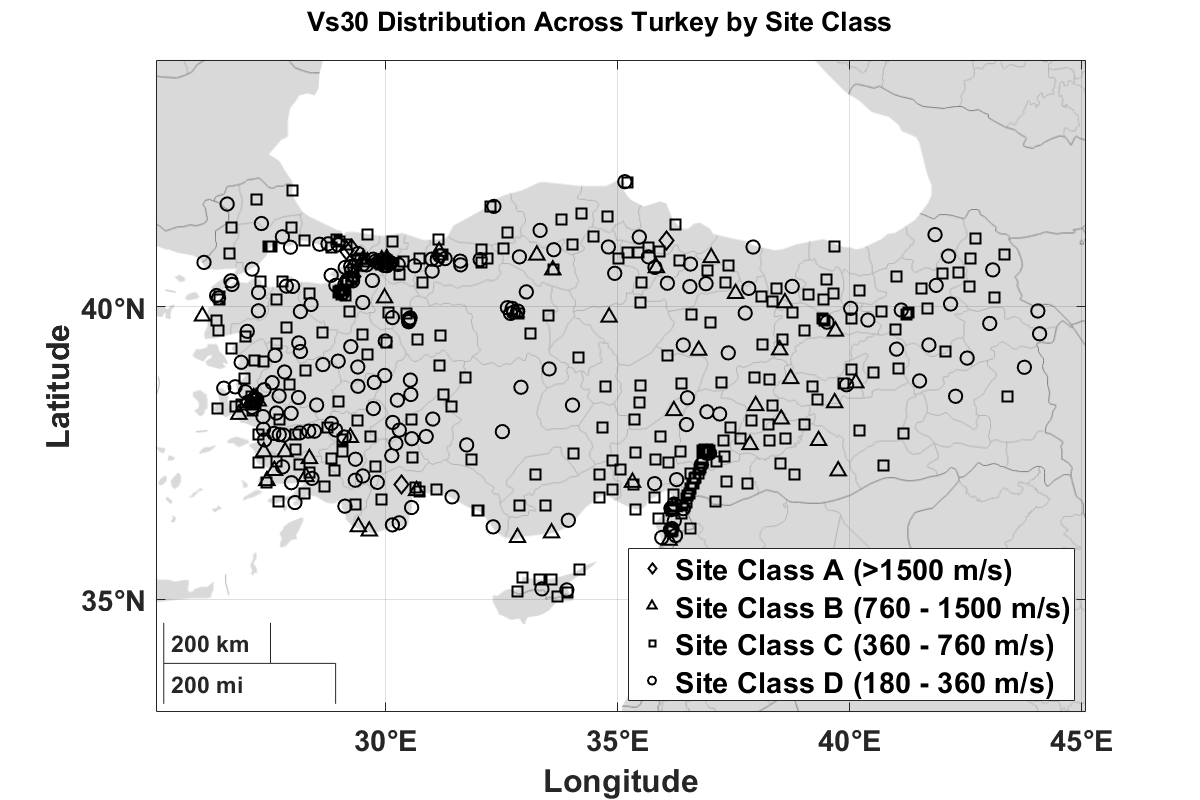} 
    \caption{Local site classes at the AFAD strong motion stations} 
    \label{fig:my_label7} 
\end{figure}  
This study is conducted using earthquake station data provided by  \cite{afad2025} in Türkiye. The dataset includes strong ground motion records from these stations, with \( V_{s30} \) values available for 594 out of the 799 strong motion stations across Türkiye and are based on MASW measurements carried out as a part of studies conducted by \citep{sandikkaya2010site} and \citep{kurtulucs2020determination}. Test reports of these tests are provided along with the publicly accessible station recordings by \cite{AFADTADAS}. 

This dataset encompasses a total of 36,417 records measured by these stations between 2012 and 2018, as detailed in the study by \cite{yilmaz2024deep}. Out of these, 13,974 records are from stations with measured $V_{s30}$ values, used as the ground truth in this study. The magnitudes of the recorded earthquakes range between M2.2 and M6.5. Figure \ref{fig:my_label7} illustrates the distribution of strong motion stations, along with their respective $V_{s30}$ site classifications based on \cite{nehrp2009} site class criteria. Among these classifications, Site Class C has the largest representation, comprising 57\% of all stations. Site Class D follows with 30.5\%, while Site Class B accounts for 11.7\%. The data indicate that recordings from stations categorized as Site Class C contribute 54.5\% of the total recordings used for $V_{s30}$ prediction. In comparison, stations classified as Site Class D provide 29.36\% of the recordings, and those in Site Class B contribute 15.79\%.

\section{Experimental Setup}
\setlength\extrarowheight{2.5pt} % Adds extra space to rows

\subsection{Deep Learning Model}
\begin{figure*}[t]
    \centering 
    \includegraphics[width=0.95\textwidth]{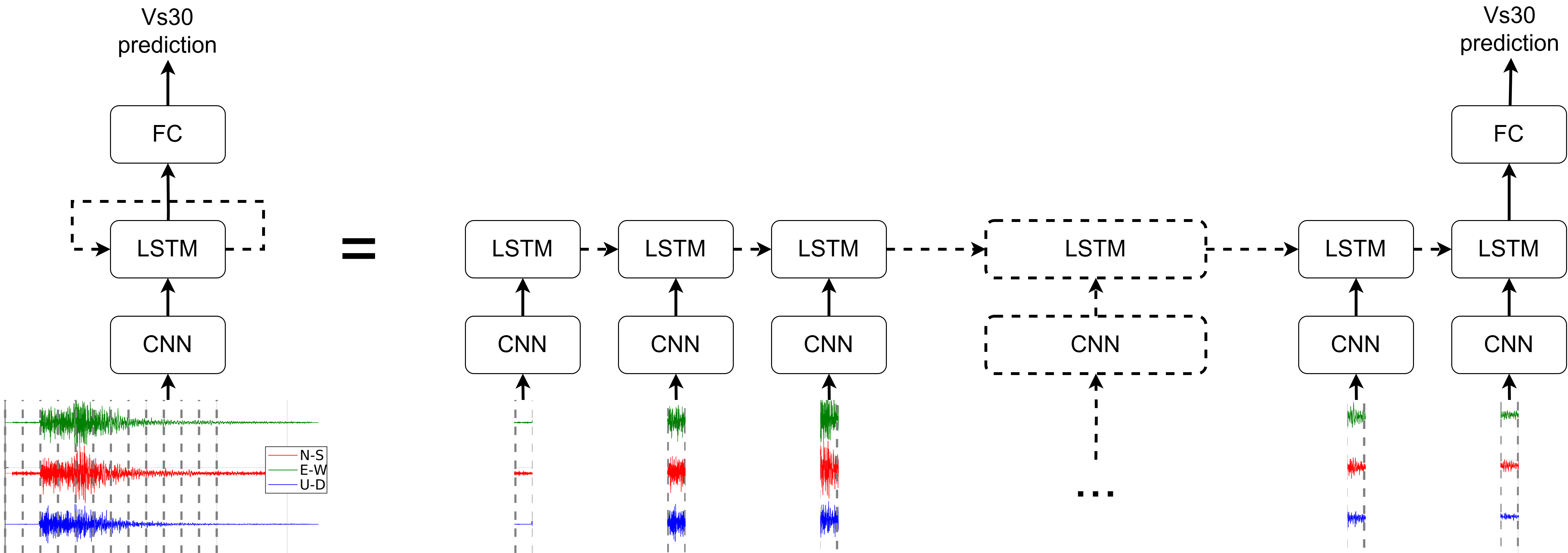} 
    \caption{Unrolled-in-time representation of the proposed CNN+LSTM sequence model.} 
    \label{fig:my_label} 
\end{figure*}

The study presented in \citep{yilmaz2024deep} demonstrated significant limitations in addressing the sequential nature of the data, as the proposed methods did not fully account for signal dependencies across time. To overcome this limitation, a CNN+LSTM-based architecture, widely utilized in the literature for sequence-based signal processing tasks, was implemented in this study, as illustrated in Figure \ref{fig:my_label}. In the proposed architecture, we aim to predict $V_{s30}$ by processing a three-channel earthquake signal, where each channel represents a different direction (East-West, North-South, Up-Down). The signal is divided into a specified number of segments (1 to 5 seconds, parameterized for each experiment). Each segment is fed into the CNN+LSTM blocks sequentially. The segments are first passed through the CNN blocks to extract features. The CNN encoder used in this study is identical to the one proposed in \citep{yilmaz2024deep}, enabling a valid ablation study to compare non-sequential and sequential models. The extracted features from each segment are then fed into the LSTM model. As new segments are fed as input, the LSTM's hidden and cell states are subsequently passed to the same LSTM block, allowing the accumulation of feature representations. The final states of the LSTM block are passed to a fully connected (FC) layer, which produces the $V_{s30}$ prediction.

\subsection{Experiments}
As shown in Table \ref{tab:my_table}, this study aims  to evaluate the impact of auxiliary information and transfer learning on the performance of deep learning models for $V_{s30}$ prediction. Our previous work \citep{ccauglar2024exploring} demonstrated that directly feeding raw accelerometer signals into deep learning models often fails to extract meaningful features, especially when auxiliary information (e.g., P/S wave arrival time information) is absent. In addition, \cite{yilmaz2024deep} highlighted the significant role of incorporating auxiliary information, such as the time of PGA, to enhance model performance. Building on these findings, we designed experiments to systematically analyze the effects of integrating auxiliary information and leveraging the CNN encoder developed in \citep{yilmaz2024deep} for transfer learning.

% \usepackage{multirow}
\begin{table}[b]
\caption{Overview of the Experiments}
\label{tab:my_table}
\centering
\begin{tabular}{|c|c|cc|c|l|}
\hline
\multirow{2}{*}{\begin{tabular}[c]{@{}c@{}}Training \\ Method\end{tabular}} & \multirow{2}{*}{\begin{tabular}[c]{@{}c@{}}P/S \\ Info\end{tabular}} & \multicolumn{2}{c|}{Annotation}                                      & \multirow{2}{*}{\begin{tabular}[c]{@{}c@{}}Signal \\ Segment\end{tabular}} & \multirow{2}{*}{Exp. Name} \\ \cline{3-4}
                                                                            &                                                                      & \multicolumn{1}{c|}{Train}                 & Test                    &                                                                            &                            \\ \hline
\multirow{8}{*}{\begin{tabular}[c]{@{}c@{}}From\\ Scratch\end{tabular}}     & \multirow{4}{*}{Yes}                                                 & \multicolumn{1}{c|}{\multirow{2}{*}{auto}} & \multirow{2}{*}{auto}   & PGA                                                                        & $\alpha_{auto,PGA}$ \\ \cline{5-6} 
                                                                            &                                                                      & \multicolumn{1}{c|}{}                      &                         & P                                                                          & $\alpha_{auto,P,15sec}$ \\ \cline{3-6} 
                                                                            &                                                                      & \multicolumn{1}{c|}{\multirow{2}{*}{auto}} & \multirow{2}{*}{manual} & PGA                                                                        & $\alpha_{man,PGA}$ \\ \cline{5-6} 
                                                                            &                                                                      & \multicolumn{1}{c|}{}                      &                         & P                                                                          & $\alpha_{man,P,15sec}$ \\ \cline{2-6} 
                                                                            & \multirow{4}{*}{No}                                                  & \multicolumn{1}{c|}{\multirow{2}{*}{auto}} & \multirow{2}{*}{auto}   & PGA                                                                        & $\beta_{auto,PGA}$  \\ \cline{5-6} 
                                                                            &                                                                      & \multicolumn{1}{c|}{}                      &                         & P                                                                          & $\beta_{auto,P,15sec}$  \\ \cline{3-6} 
                                                                            &                                                                      & \multicolumn{1}{c|}{\multirow{2}{*}{auto}} & \multirow{2}{*}{manual} & PGA                                                                        & $\beta_{man,PGA}$  \\ \cline{5-6} 
                                                                            &                                                                      & \multicolumn{1}{c|}{}                      &                         & P                                                                          & $\beta_{man,P,15sec}$  \\ \hline
\multirow{4}{*}{Transfer}                                                   & \multirow{2}{*}{Yes}                                                 & \multicolumn{1}{c|}{auto}                  & auto                    & PGA                                                                        &  $\gamma_{ps,auto}$ \\ \cline{3-6} 
                                                                            &                                                                      & \multicolumn{1}{c|}{auto}                  & manual                  & PGA                                                                        & $\gamma_{ps,man}$ \\ \cline{2-6} 
                                                                            & \multirow{2}{*}{No}                                                  & \multicolumn{1}{c|}{auto}                  & auto                    & PGA                                                                        &  $\gamma_{-,auto}$ \\ \cline{3-6} 
                                                                            &                                                                      & \multicolumn{1}{c|}{auto}                  & manual                  & PGA                                                                        &  $\gamma_{-,manl}$ \\ \hline
\end{tabular}
\end{table}

\begin{figure}[t]
    \centering 
    \includegraphics[width=0.49\textwidth, trim=0 0 0 20, clip]{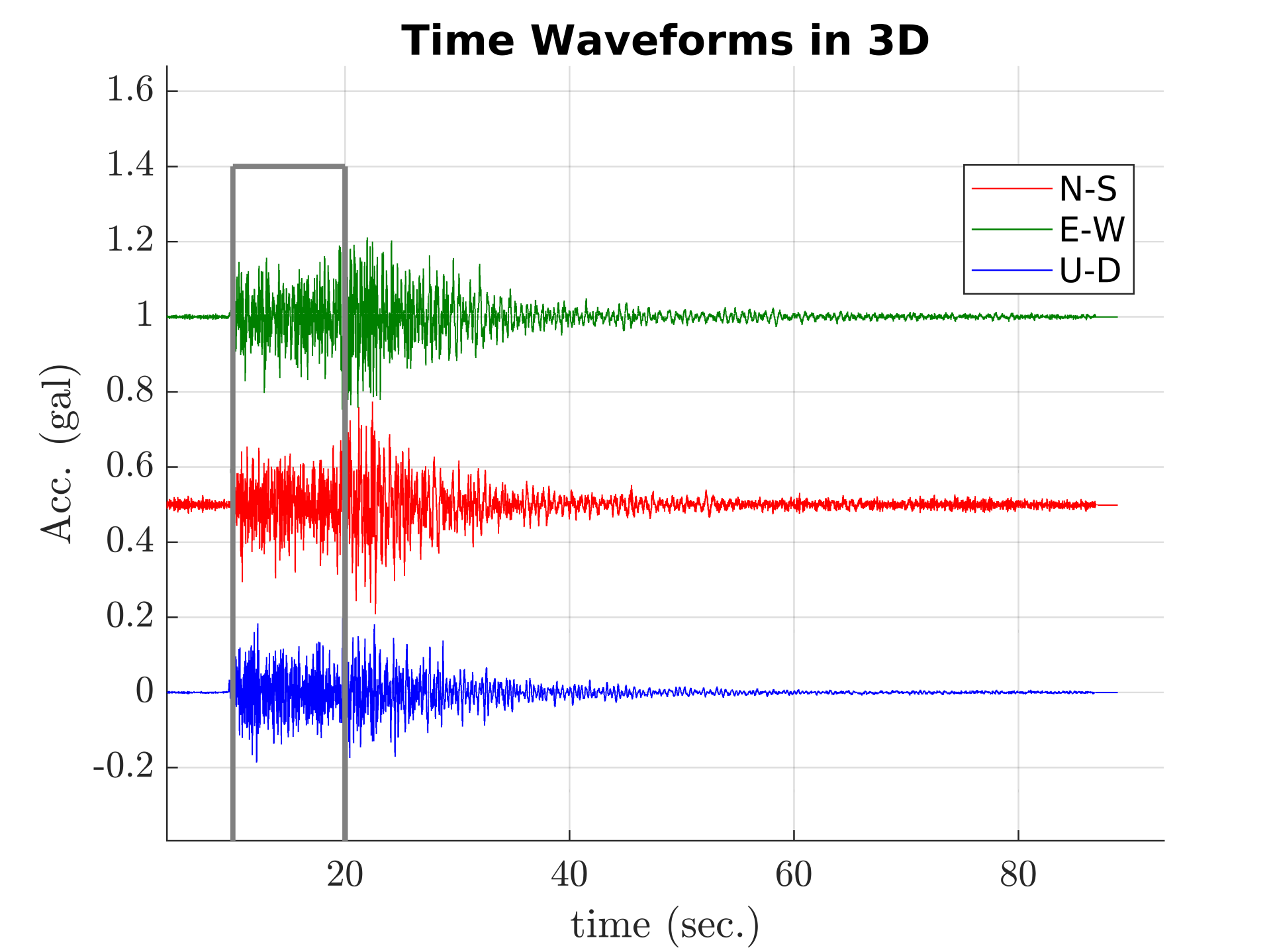} 
   
    \caption{A 4-channel input structure highlighting the arrival times of P and S waves} 
    \label{fig:my_label2} 
\end{figure}

Table \ref{tab:my_table} outlines the various experimental configurations created to evaluate these effects. The experiments are categorized as follows:
\begin{itemize}
    \item \emph{Training Strategy}: Two primary strategies are compared—training the model from scratch and transferring the CNN encoder from \cite{yilmaz2024deep}.
    \item \emph{Auxiliary P/S Information}: The experiments include configurations where P/S wave arrival time information was either provided as an additional 4$^{th}$ input channel (Figure \ref{fig:my_label2}) or not. 
    \item \emph{Annotation}: For experiments utilizing P/S wave information, the annotations are either automatically generated (noisy labels) or manually labeled by experts (hard labels), as indicated in the ``Annotation'' column.
    \item \emph{Signal Segmentation}: The entire input signal (that the sequential segments are sampled from) is either centered around the P-wave arrival time or the PGA time (as in \cite{yilmaz2024deep}), as specified in the ``Signal Segment'' column. When selecting segments centered around the P-wave arrival, the P-wave arrival time would always be near the beginning of the earthquake signal in cases where 60-second signals are used, preventing a focused representation of the P-wave itself. To address this, we reduced the length of the signals from 60 seconds to 15 seconds, ensuring that the P-wave arrival serves as the midpoint of the signal. This approach allowed us to focus on the P-wave arrival while retaining the critical portions of the signal.

\end{itemize}

As indicated in Table \ref{tab:my_table} column ``P/S Info'', following the approach proposed by \cite{Mousavi2020a}, experiments were conducted with an additional input channel that utilized P/S arrival times, as shown in Figure \ref{fig:my_label2}. In this way, we aimed to help the model make better sense of the signal by highlighting the part of the signal from the arrival moment the P wave to the arrival moment of the S wave. The main challenge in using P/S arrival times as input for the entire dataset is the lack of available or annotated information on P and S arrival times for large dataset of accelerometer recordings. Various annotation methods were tested for this purpose. 

The first method involved manual annotation using an interface we developed. The second method employed an algorithm proposed by \cite{kalkan2016automatic}, which enables the annotation of P- and S-wave arrivals along with their corresponding signal-to-noise ratio (SNR). While the first method proved reliable in terms of accuracy, it was only feasible for a limited dataset of 2000 records due to the large volume of accelerometer data. The second method, however, did not meet the desired reliability standards and was therefore not used. Under these circumstances, experiments were conducted with the limited dataset where P- and S-wave arrivals were manually annotated. As expected, the limited data size prevented achieving the desired results. Based on these findings, a third method, namely the EQTransformer (EQT), a deep learning approach based on LSTMs and Transformers developed by \cite{Mousavi2020b}, was tested. The reliability of the results and the significantly larger dataset made this method the preferred choice for this study. 
%To annotate P and S arrivals in our accelerometer data using the EQT method, we utilized QuakeLabeler \cite{mai2022quakelabeler} for format conversion. 
Consequently, we obtained a set consisting of 11,840 records annotated using the EQT method. 

%Based on the availability of this additional information, the training and testing datasets were reorganized to increase the diversity of the experiments. 

We experimented with identifying critical regions of the accelerometer signals to evaluate their impact on $V_{s30}$ prediction. Specifically, signals centered around the peak ground acceleration region were segmented into smaller parts and sequentially fed into the deep learning model. %The motivation for selecting the PGA region is that it represents the signal's peak acceleration component, which is likely to carry feature-rich information.
Our objective to center the window around the PGA is to ensure a consistent representation across all signals in the dataset. 
However, focusing on the PGA region does not mean excluding the P- and S-wave phases where site related characteristics are likely present as well; instead, a significant portion of the signal is still retained. Furthermore, to investigate how different time windows affect the results, we selected a region around the P-wave arrival time at the station. Similar to the PGA approach, the selected region around the P-wave was segmented into smaller parts before being fed into the model.

The rationale for choosing the P-wave region is based on previous studies in the literature \citep{kim2016subsurface}; \citep{ni2014estimating}; \citep{li2023evaluation}, which demonstrated the potential of $V_{s30}$ estimation methods based on earthquake records without applying artificial intelligence methods. Those studies primarily relied on features extracted from the signal near the P-wave arrival time. In our experiments, we aimed to assess the performance of models trained on segments derived from both regions (PGA and P-wave) and to evaluate the relative importance of the P-wave arrival time for $V_{s30}$ estimation.

\subsection{Sequence Input}

In our experiments, accelerometer data paired with $V_{s30}$ ground truth values were divided into equal-sized segments, with each segment being one second long. Each segment was fed into identical CNN layers, which were designed using the same CNN architecture as our previous work \citep{yilmaz2024deep}, allowing spatial feature extraction. The features extracted by the CNNs were then passed sequentially to LSTM layers. In this step, the LSTM layers accumulated temporal information by retaining hidden and cell states across segments. Finally, a fully connected layer was used to predict the $V_{s30}$ values. This  approach is designed to ensure that spatial features are effectively extracted by the CNN, the temporal dependencies are captured by the LSTM, and the final predictions are made using the combined temporal features.

\section{Results}
% \usepackage{multirow}

We provided an overview of all experiments, including their configuration details, previously in Table \ref{tab:my_table}. In Table \ref{tab:my_table2}, we examine five selected experiments (namely: {\textbf{$\beta_{auto,PGA}$}}, {\textbf{$\alpha_{man,PGA}$}},  {\textbf{$\alpha_{auto,P,15sec}$}}, {\textbf{$\alpha_{auto,PGA}$}}, and {\textbf{$\gamma_{ps,auto}$}}) in greater detail, focusing on parameters that reflect a general consensus and significantly influence the results. Additionally, the selected results presented in Table \ref{tab:my_table2} represent the test set outcomes, and the station counts correspond to the number of stations in the test set data. Specifically, we aim to investigate the impact of adding P/S information on the outcomes, the effect of different annotation methods for P/S information, and the influence of selecting signals around the P-wave or PGA regions. Additionally in this table,  we observe how these parameters perform across different site classes, represented by $V_{s30}$ ranges.

We conducted twelve experiments to explore a variety of parameter configurations and understand the factors affecting $V_{s30}$ predictions. 
The five experiments detailed in Table \ref{tab:my_table2}) sufficiently represent the trends observed in all experiments. %The selected experiments were carefully chosen because they summarize the key findings across different parameter variations, including the use of P/S information, annotation methods, and signal regions (e.g., around the P-wave or PGA).%

% \usepackage{multirow}
% \usepackage{multirow}
% \usepackage{multirow}
\begin{table}[t]
\caption{Results w.r.t. site classes.} %{\textbf{$\beta_{auto,PGA}$}}: Average error categorized by site classes of training from scratch experiment, {\textbf{$\alpha_{man,PGA}$}}: Average error categorized by site classes of manually annotated P/S added experiment, {\textbf{$\alpha_{auto,P,15sec}$}}: Average error categorized by site classes of EQT annotated P/S added experiment (15 second long signal), {\textbf{$\alpha_{auto,PGA}$}}: Average error categorized by site classes of EQT annotated P/S added experiment (segments selected around the arrival time of the P-wave), {\textbf{$\gamma_{ps,auto}$}}: Average error categorized by site classes of transfer learning experiment.}
\label{tab:my_table2}
% \usepackage{multirow}
\begin{tabular}{|l|crr|}
\hline
\textbf{Experiment}                           & \multicolumn{1}{c|}{\textbf{Site Class}} & \multicolumn{1}{c}{\textbf{\begin{tabular}[c]{@{}c@{}}No. of \\ Stations\end{tabular}}} & \multicolumn{1}{c|}{\textbf{Absolute Mean Error}} \\ \hline
\multirow{5}{*}{\textbf{$\beta_{auto,PGA}$}}  & \multicolumn{1}{c|}{A}                   & 3                                                                                       & 70.53\%                                           \\
                                              & \multicolumn{1}{c|}{B}                   & 15                                                                                      & 45.57\%                                           \\
                                              & \multicolumn{1}{c|}{C}                   & 81                                                                                      & 17.77\%                                           \\
                                              & \multicolumn{1}{c|}{D}                   & 21                                                                                      & 71.01\%                                           \\ \cline{2-4} 
                                              & \textbf{Total}                           & \textbf{120}                                                                            & \textbf{37.84\%}                                  \\ \hline
\multirow{5}{*}{\textbf{$\alpha_{man,PGA}$}}  & \multicolumn{1}{c|}{A}                   & 0                                                                                       & NaN                                               \\
                                              & \multicolumn{1}{c|}{B}                   & 7                                                                                       & 41.24\%                                           \\
                                              & \multicolumn{1}{c|}{C}                   & 39                                                                                      & 18.15\%                                           \\
                                              & \multicolumn{1}{c|}{D}                   & 6                                                                                       & 67.10\%                                           \\ \cline{2-4} 
                                              & \textbf{Total}                           & \textbf{52}                                                                             & \textbf{25.03\%}                                  \\ \hline
\multirow{5}{*}{\textbf{$\alpha_{auto,P,15sec}$}}   & \multicolumn{1}{c|}{A}                   & 0                                                                                       & NaN                                               \\
                                              & \multicolumn{1}{c|}{B}                   & 19                                                                                      & 46.49\%                                           \\
                                              & \multicolumn{1}{c|}{C}                   & 100                                                                                      & 21.54\%                                           \\
                                              & \multicolumn{1}{c|}{D}                   & 27                                                                                      & 73.16\%                                           \\ \cline{2-4} 
                                              & \textbf{Total}                           & \textbf{147}                                                                             & \textbf{38.24\%}                                  \\ \hline
\multirow{5}{*}{\textbf{$\alpha_{auto,PGA}$}} & \multicolumn{1}{c|}{A}                   & 2                                                                                       & 74.13\%                                           \\
                                              & \multicolumn{1}{c|}{B}                   & 19                                                                                      & 52.36\%                                           \\
                                              & \multicolumn{1}{c|}{C}                   & 93                                                                                      & 18.09\%                                           \\
                                              & \multicolumn{1}{c|}{D}                   & 36                                                                                      & 54.69\%                                           \\ \cline{2-4} 
                                              & \textbf{Total}                           & \textbf{150}                                                                            & \textbf{35.70\%}                                  \\ \hline
\multirow{5}{*}{\textbf{$\gamma_{ps,auto}$}}  & \multicolumn{1}{c|}{A}                   & 3                                                                                       & 79.21\%                                           \\
                                              & \multicolumn{1}{c|}{B}                   & 15                                                                                      & 63.14\%                                           \\
                                              & \multicolumn{1}{c|}{C}                   & 92                                                                                      & 29.09\%                                           \\
                                              & \multicolumn{1}{c|}{D}                   & 37                                                                                      & 15.41\%                                           \\ \cline{2-4} 
                                              & \textbf{Total}                           & \textbf{147}                                                                            & \textbf{34.63\%}                                  \\ \hline
\end{tabular}
\end{table}
\begin{figure*}[p]
    \centering
    % Experiment Alpha
    \begin{subfigure}[b]{1.0\textwidth}
        \centering
        \includegraphics[width=0.32\textwidth]{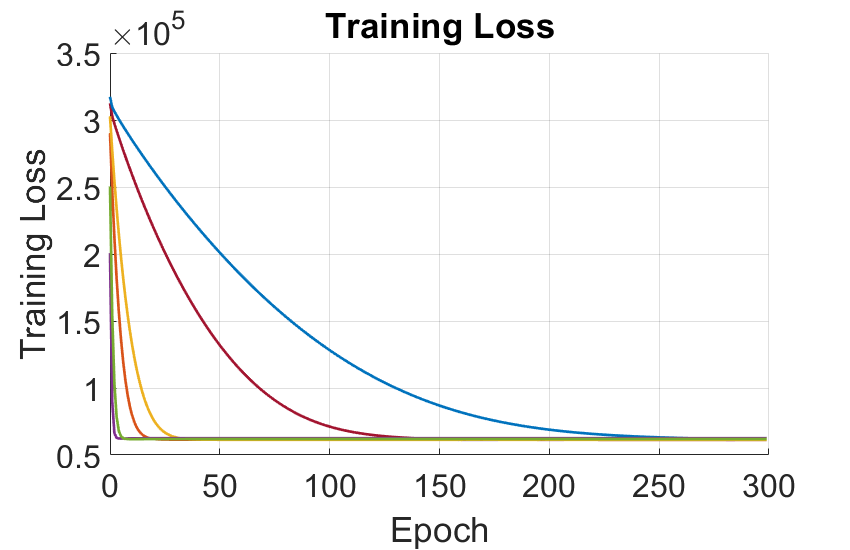}
        %\caption{Training loss graph of training from scratch experiment}
        %\label{fig:training_loss}
    %\end{subfigure}
    %\hfill
    %\begin{subfigure}[b]{0.26\textwidth}
    %    \centering
        \includegraphics[width=0.32\textwidth]{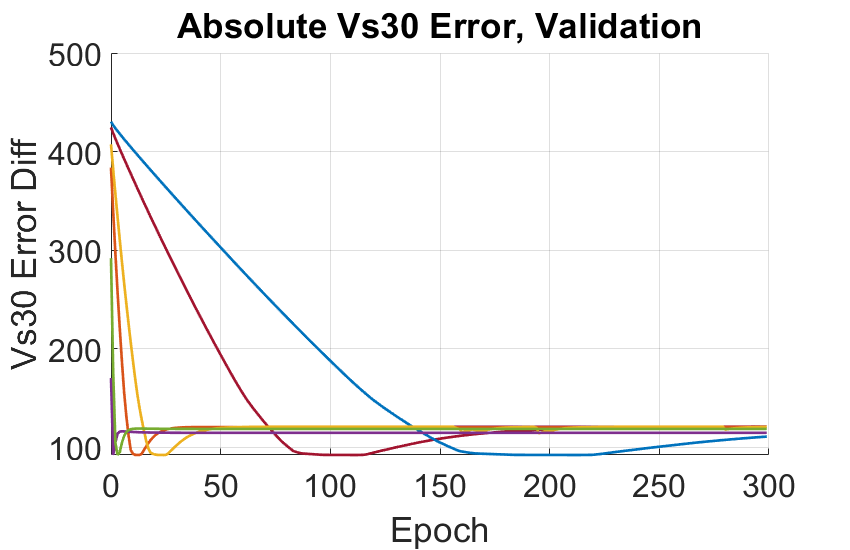}
    %    \caption{$V_{s30}$ difference graph of training from scratch experiment}
    %    \label{fig:$V_{s30}$_diff}
    %\end{subfigure}
    %\hfill
    %\begin{subfigure}[b]{0.26\textwidth}
    %    \centering
        \includegraphics[width=0.32\textwidth]{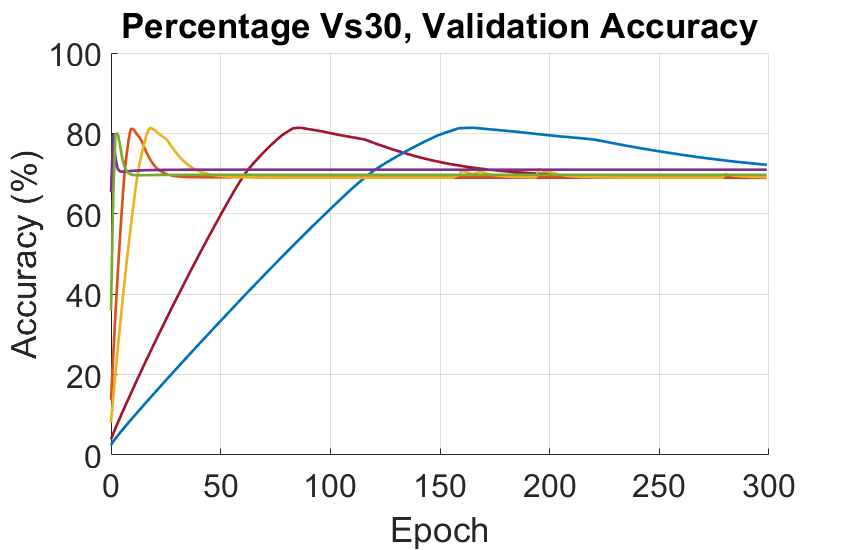}
   %     \caption{$V_{s30}$ percentage difference graph of training from scratch experiment}
   %     \label{fig:$V_{s30}$_percentage_diff}

        \caption{{\textbf{$\beta_{auto,PGA}$}}}
        \label{fig:beta-auto-PGA}
    \end{subfigure}

    % Experiment Beta
    \begin{subfigure}[b]{1.0\textwidth}
        \centering
        \includegraphics[width=0.32\textwidth]{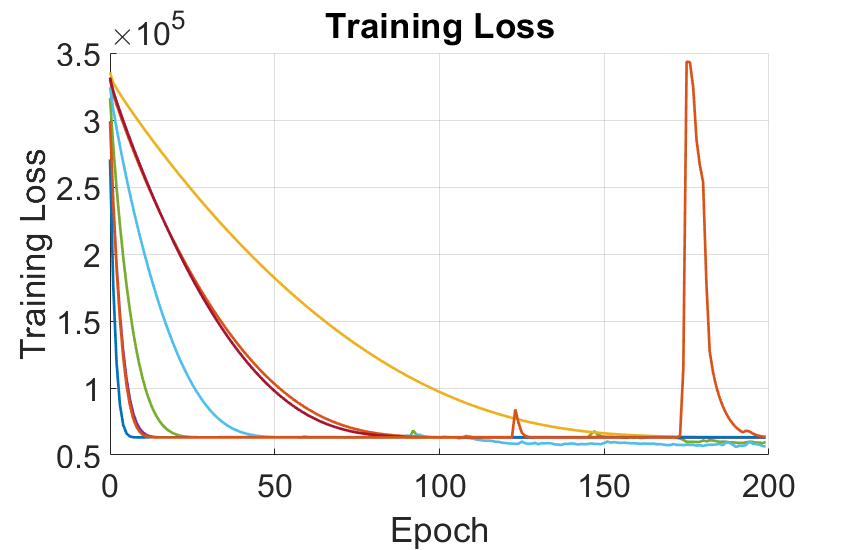}
        %\caption{Training loss graph of manually annotated P/S added experiment}
        %\label{fig:training_loss_ps}
    %\end{subfigure}
    %\hfill
    %\begin{subfigure}[b]{0.26\textwidth}
        %\centering
        \includegraphics[width=0.32\textwidth]{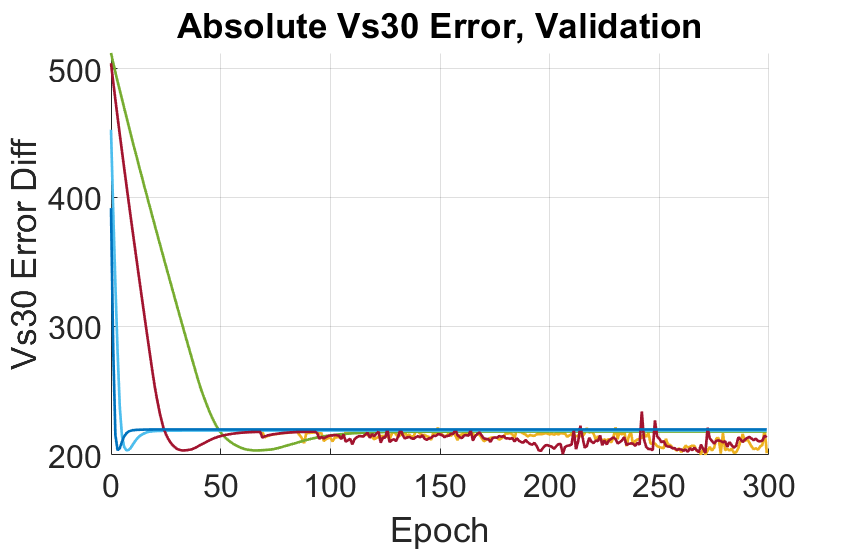}
        %\caption{$V_{s30}$ difference graph of manually annotated P/S added experiment}
        %\label{fig:$V_{s30}$_diff_ps}
    %\end{subfigure}
    %\hfill
    %\begin{subfigure}[b]{0.26\textwidth}
        %\centering
        \includegraphics[width=0.32\textwidth]{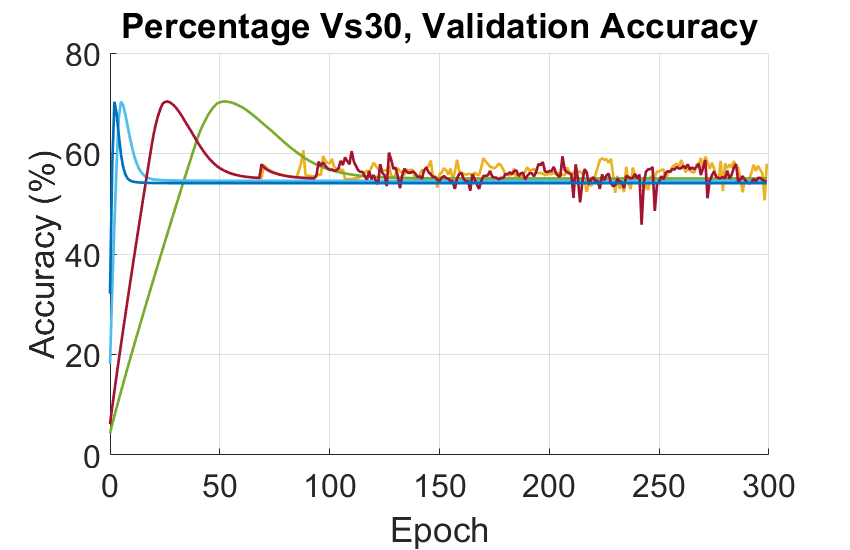}
        %\caption{$V_{s30}$ percentage difference graph of manually annotated P/S added experiment}
        %\label{fig:$V_{s30}$_percentage_diff_ps}
        \caption{{\textbf{$\alpha_{man,PGA}$}}}
        \label{fig:alpha-man-PGA}
    \end{subfigure}

    % Experiment Gamma
    \begin{subfigure}[b]{1.0\textwidth}
        \centering
        \includegraphics[width=0.32\textwidth]{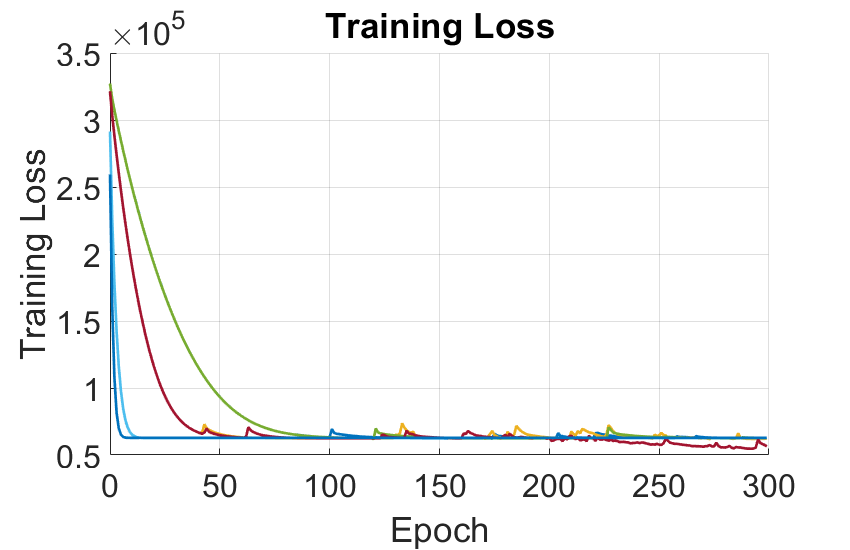}
        %\caption{Training loss graph of EQT annotated P/S added experiment.}
        %\label{fig:training_loss_eqt}
    %\end{subfigure}
    %\hfill
    %\begin{subfigure}[b]{0.26\textwidth}
        %\centering
        \includegraphics[width=0.32\textwidth]{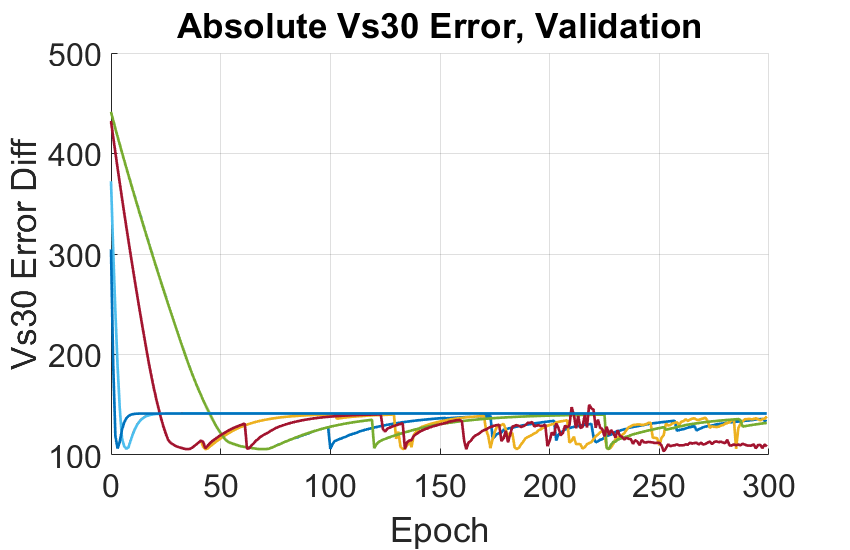}
        %\caption{$V_{s30}$ difference graph of EQT annotated P/S added experiment.}
        %\label{fig:$V_{s30}$_diff_eqt}
    %\end{subfigure}
    %\hfill
    %\begin{subfigure}[b]{0.26\textwidth}
        %\centering
        \includegraphics[width=0.32\textwidth]{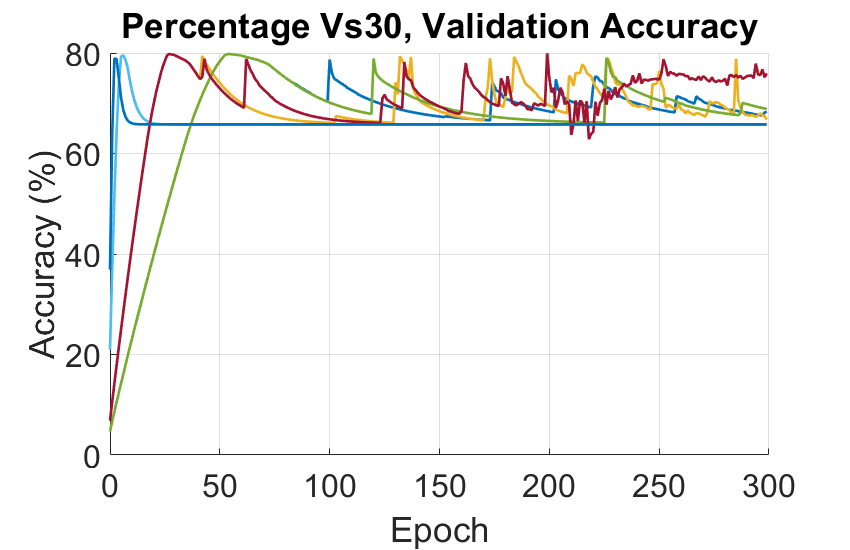}
        %\caption{$V_{s30}$ percentage difference graph of EQT annotated P/S added experiment.}
        %\label{fig:$V_{s30}$_percentage_diff_eqt}
        \caption{        {\textbf{$\alpha_{auto,P,15sec}$}}}
        \label{fig:alpha-auto-p-15}
    \end{subfigure}

    % Experiment Omega
    \begin{subfigure}[b]{1.0\textwidth}
        \centering
        \includegraphics[width=0.32\textwidth]{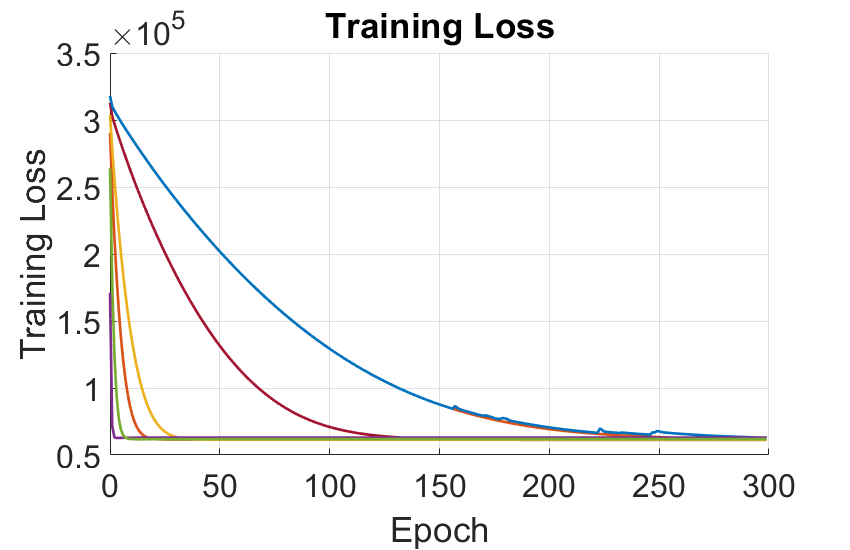}
        %\caption{Training loss graph of P-wave centric experiment.}
        %\label{fig:training_loss_pwave}
    %\end{subfigure}
    %\hfill
    %\begin{subfigure}[b]{0.26\textwidth}
        %\centering
        \includegraphics[width=0.32\textwidth]{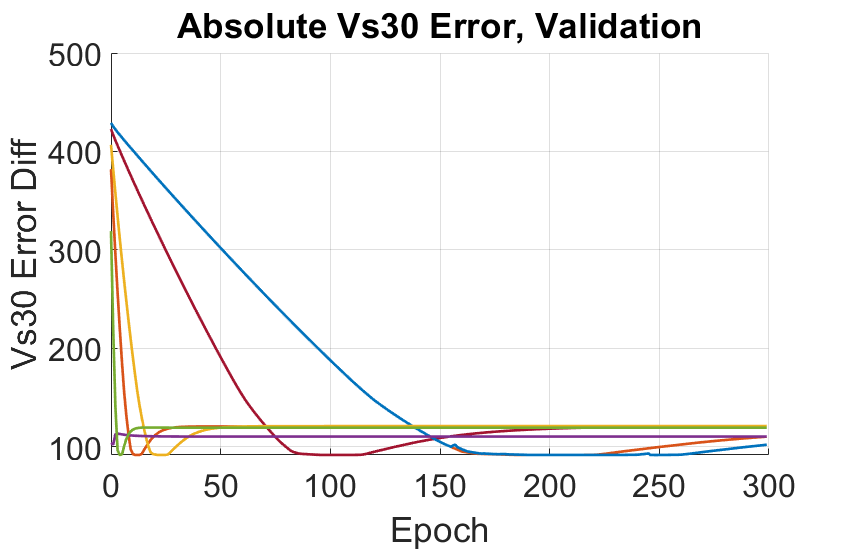}
        %\caption{$V_{s30}$ difference graph of P-wave centric experiment.}
        %\label{fig:$V_{s30}$_diff_pwave}
    %\end{subfigure}
    %\hfill
    %\begin{subfigure}[b]{0.26\textwidth}
        %\centering
        \includegraphics[width=0.32\textwidth]{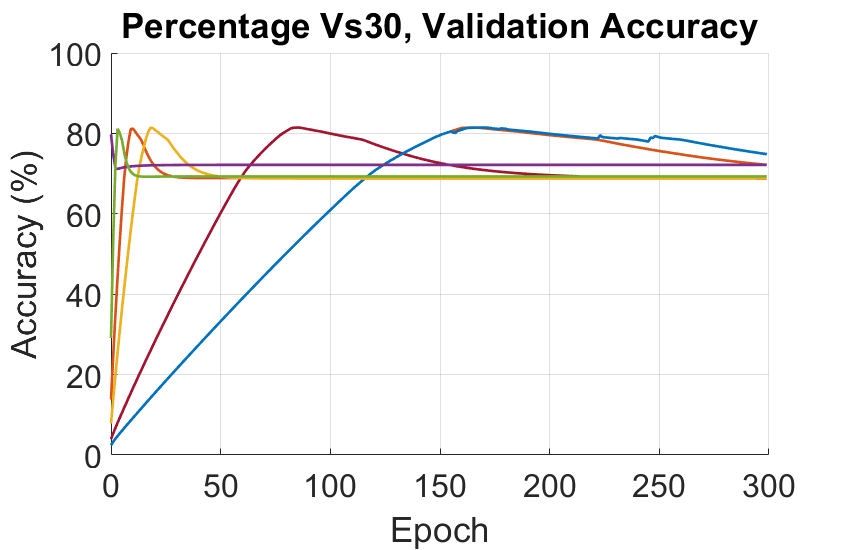}
        %\caption{$V_{s30}$ percentage difference graph of P-wave centric experiment.}
        %\label{fig:$V_{s30}$_percentage_diff_pwave}
        \caption{{\textbf{$\alpha_{auto,PGA}$}}}
        \label{fig:alpha-auto-PGA}
    \end{subfigure}

    % Experiment Delta
    \begin{subfigure}[b]{1.0\textwidth}
        \centering
        \includegraphics[width=0.32\textwidth]{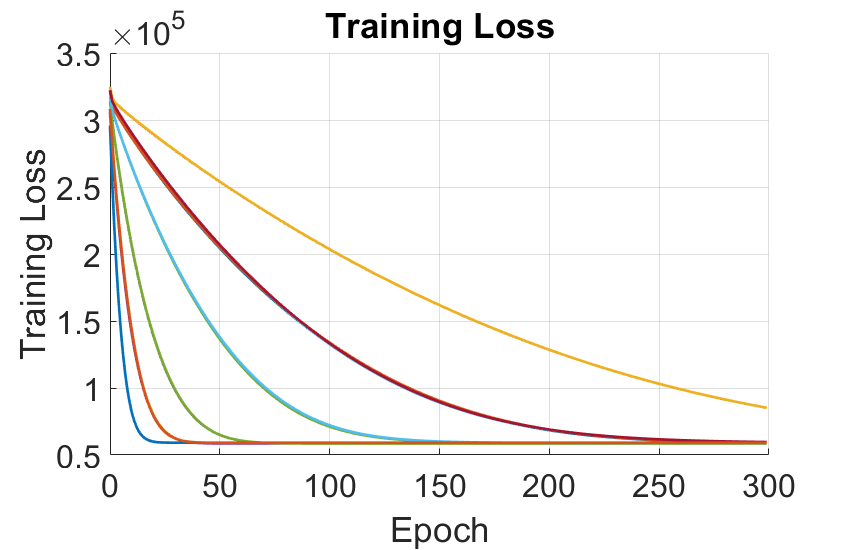}
        %\caption{Training loss graph of transfer learning experiment.}
        %\label{fig:training_loss_transfer}
    %\end{subfigure}
    %\hfill
    %\begin{subfigure}[b]{0.26\textwidth}
        %\centering
        \includegraphics[width=0.32\textwidth]{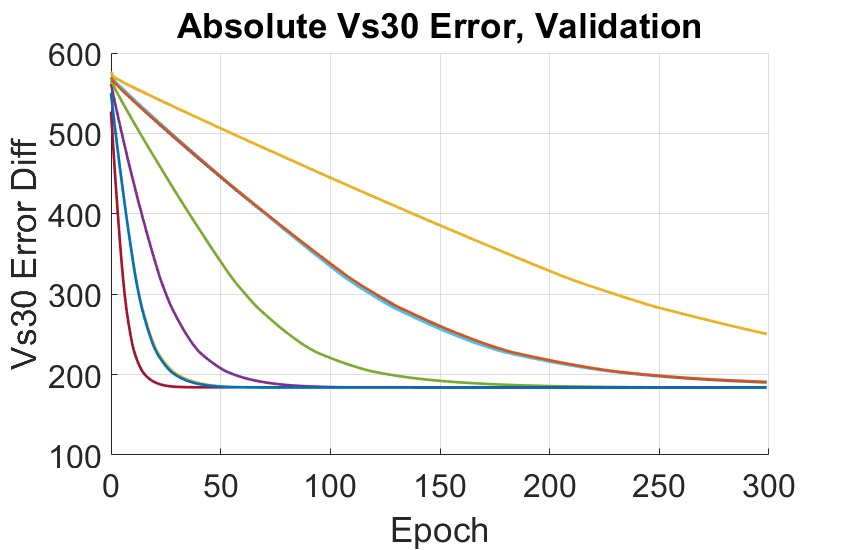}
        %\caption{$V_{s30}$ difference graph of transfer learning experiment.}
        %\label{fig:$V_{s30}$_diff_transfer}
    %\end{subfigure}
    %\hfill
    %\begin{subfigure}[b]{0.26\textwidth}
        %\centering
        \includegraphics[width=0.32\textwidth]{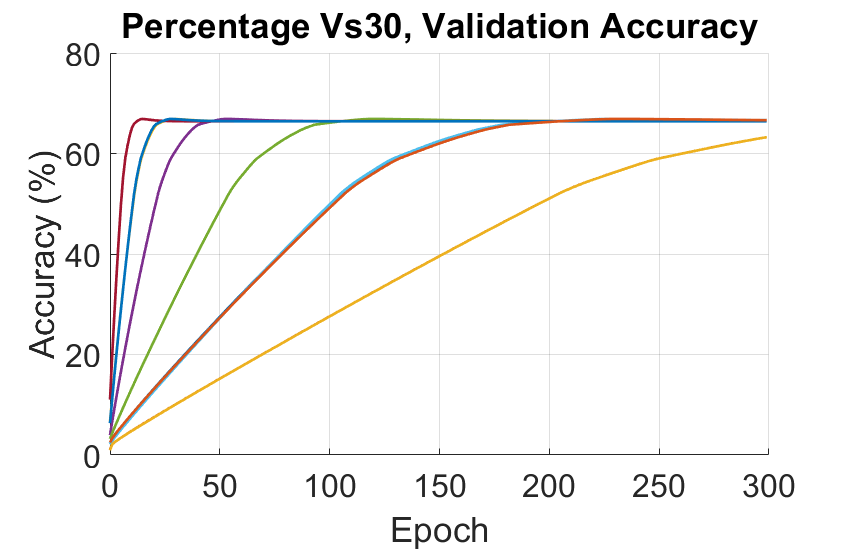}
        %\caption{$V_{s30}$ percentage difference graph of transfer learning experiment.}
        %\label{fig:$V_{s30}$_percentage_diff_transfer}
        \caption{{\textbf{$\gamma_{ps,auto}$}}}
        \label{fig:gama-ps-auto}
        
    \end{subfigure}

    \caption{Visualizations of the training losses (left-most column), $V_{s30}$ prediction errors (middle column) and $V_{s30}$ percentage prediction error (right column) of experiments (from top to bottom): a) {\textbf{$\beta_{auto,PGA}$}}, b) {\textbf{$\alpha_{man,PGA}$}},  c) {\textbf{$\alpha_{auto,P,15sec}$}},  d) {\textbf{$\alpha_{auto,PGA}$}}, and e) {\textbf{$\gamma_{ps,auto}$}}. All experiments are repeated with varying hyperparameters. The bold line represents the average of the results, with the transparent shaded regions indicating the maximum and minimum ranges.}
    \label{fig:grouped_figures_all}
\end{figure*}

Figure \ref{fig:grouped_figures_all} presents the graphical results of five selected experiments, which best represent the overall findings. For each experiment, three different graphs are provided: the training Loss, the difference between predicted and actual $V_{s30}$ values in the validation set, and the percentage accuracy of $V_{s30}$ predictions in the validation set. We selected these visualizations to evaluate the training and validation processes, ensuring that the test results in Table \ref{tab:my_table2} come from well-trained models.

The results presented in Figure \ref{fig:beta-auto-PGA} are from our baseline experiment $\beta_{auto,PGA}$, performed without adding any P/S wave information, shown as the first of the five in Table \ref{tab:my_table2}.

In the following, the graphs of the experiment $\alpha_{man,PGA}$ performed by adding P/S arrival times are presented in Figure \ref{fig:alpha-man-PGA}. The P/S values here were manually annotated by an expert. Hence for this model, we train the model using the existing P/S-annotated data. The third experiment we focus on is namely $\alpha_{auto,P,15sec}$, for which the P/S information was automatically annotated using the EQTransformer \citep{Mousavi2020b} model. The results of this experiment is shown in Figure \ref{fig:alpha-auto-p-15}. Results of the experiment $\alpha_{auto,PGA}$, which uses the accelerometer data segments selected around the arrival time of the PGA , instead of the P-wave, are presented in Figure \ref{fig:alpha-auto-PGA}.

In the fifth experiment that we focus our analysis, namely $\gamma_{ps,auto}$, instead of training the encoder from scratch, we transfer encoder from our previous study by \cite{yilmaz2024deep}. By adjusting the learned knowledge, we aim to get more specific and customized results. Unsurprisingly, this approach sped up the learning process and improved computational performance. In this transfer experiment, as in the previous stage, two different approaches were tested. In the first approach, the dataset with 11,840 automatically P-S annotated data points was used as the training set, while the manually annotated data served as the test set. In the second approach, the training, validation, and test sets were all selected from the automatically annotated dataset of 11,840 accelerometer records and fed into the CNN+LSTM architecture. The results presented in \ref{tab:my_table2} correspond to the automatically annotated dataset, as this approach yielded better performance. The results from the experiments are shown in Figure \ref{fig:gama-ps-auto}. \\

\section{Discussions}

In the following section, we address several key comparisons. We examine how factors such as experiments that incorporated P/S wave arrival information and the use of specific segments of the analyzed signal influenced the results. We also analyze the observed accuracy trends and fluctuations, discussing the potential reasons behind these patterns. Additionally, we conduct a regional analysis to highlight the impact of geographical locations.

Table \ref{tab:my_table2} presents the best results obtained from five different experiments with optimal hyperparameters. As observed in the graphs, accuracy first increases as expected but then fluctuates before eventually stabilizing. Also, validation accuracy is consistently higher than training accuracy, though both eventually level off at the same accuracy. % The training graphs indicate that the model begins overfitting at certain points. 
At those values, training was stopped, and test results were recorded. Therefore, the final reported results correspond to the best-performing stage. %before overfitting occurred, rather than the last stage of training. 
%Early in training, the model learns general patterns in the data, improving accuracy. But as training continues, it starts memorizing the training set instead of generalizing, making it less effective on new data. This could explain why accuracy drops after reaching an initial peak. We addressed this problem by obtaining our test results from the point at which learning transitioned to memorization and during the experiments.

Considering all our experiments, we can conclude that the best results were obtained in our transfer learning experiment. It achieved the lowest error rate in percentage terms, and the error rates across site classes were more evenly distributed. Comparing the $\beta_{auto,PGA}$ and $\alpha_{auto,PGA}$ experiments, we observe that adding only the P-wave arrival time has a positive impact on the results. We also found that using manually annotated data as the test set ($\alpha_{man,PGA}$) (Figure \ref{fig:alpha-man-PGA}) led to higher accuracy, likely due to the smaller dataset size introducing bias. Additionally, the $\gamma_{ps,auto}$ experiment, which achieved the best results among the automatically annotated cases, shows that the earthquake-related feature weights obtained in our previous study \citep{yilmaz2024deep} were effective when combined with the additional information used in this study In addition to results of Site Class C, which consistently outperformed other site classes in most experiments. The results for Site Class D also showed promising improvements in transfer learning experiment. The non-sequential ResNet \citep{he2016deep} and TCN \citep{Mousavi2020a} architectures, which we used in our previous study \citep{yilmaz2024deep}, maintained a more balanced learning process by capturing both local and global patterns. This led to better generalization across different $V_{s30}$ distributions. Consequently, the results of the experiments conducted with the weights transferred from our previous study were significantly better compared to the others.

Overall, we achieved better predictions for the most frequent $V_{s30}$ values (Site Class C), with accuracy improving as the proportion of these stations increased. The inclusion of P/S information contributed positively across all experiments, and the transfer learning approach proved successful due to the effective features derived from our prior work. Also, the transfer learning experiment ($\gamma_{ps,auto}$) (Figure \ref{fig:gama-ps-auto}) outperformed the baseline experiment ($\beta_{auto,PGA}$) (Figure \ref{fig:beta-auto-PGA}), confirming that insights from our previous work were effective for generalization.

In light of these successful results, it is important to note that the CNN+LSTM model encountered challenges in generalization, possibly due to its higher number of parameters and increased sensitivity to overfitting. While the LSTM layer is designed to capture long-term dependencies, it may have focused too much on repetitive patterns instead of learning generalizable features. Data imbalance is also a potential issue. Some \( V_{s30} \) values appear much more often than others in the dataset. In particular, Site Class C has significantly more samples, creating an uneven distribution. As a result, the model struggles to make accurate predictions for rare $V_{s30}$ values, leading to lower performance on those cases.

\subsection{Regional Analysis}
We conducted a regional analysis to determine whether a regional distinction can be made in the prediction of \( V_{s30} \) values across different geographical areas and to observe their performance. 
%First, we sought to determine whether accelerometer signals alone could provide sufficient features for accurate $V_{s30}$ predictions, without relying on supplementary geophysical data. Second, we explored whether incorporating additional information, such as P-wave and S-wave arrival times, alongside the input signal could enhance the model's feature extraction capabilities. Lastly, we examined the impact of focusing on specific regions of the signal, such as segments around the PGA or the P-wave arrival time, to assess whether these critical regions provide a performance advantage in the prediction task. These hypotheses were central to guiding our approach and provided the foundation for interpreting the experimental results presented in the relevant sections.

\begin{figure}[t]
    \centering 
    \includegraphics[width=0.5\textwidth, trim=50 0 50 0, clip]{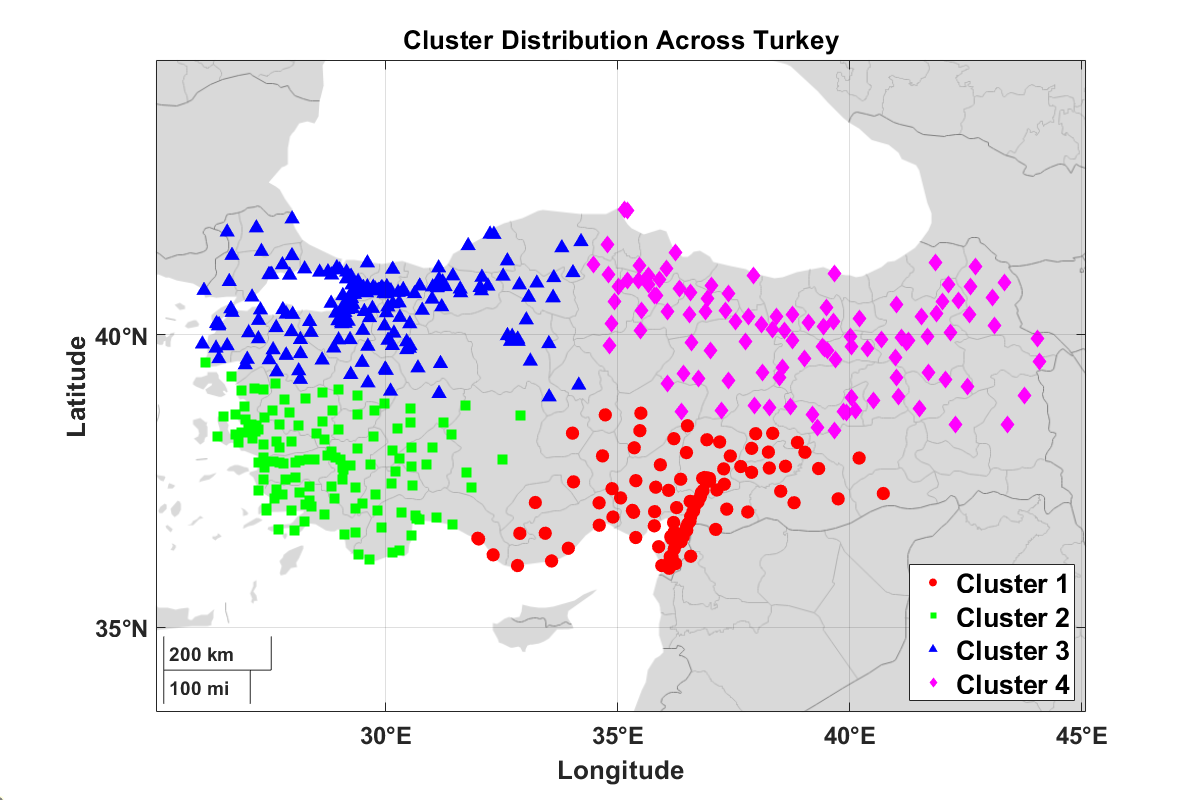} 
    \caption{Stations with $V_{s30}$ data measured according to the K-means clustering method (585) stations divided into clusters (regions) (colors represent different clusters).} 
    \label{fig:my_label4} 
\end{figure}

In order to further evaluate all these results, regional average logarithmic error histogram were created. While conducting this histogram and analysis, we utilized the results of transfer experiment ({\textbf{$\gamma_{ps,auto}$}}), which we consider to be the best among them, as discussed in the discussions section. For this purpose, stations were clustered according to their location, geology and the lithology at the station location using the k-means algorithm \citep{macqueen1967some}.  The optimum number of clusters was selected using the elbow method \citep{tibshirani2001estimating}.  As a result of this process, 4 clusters (regions) were obtained (Figure \ref{fig:my_label4}). In our previous study \cite{turkmen2024deep}, which aimed at epicenter localization, we identified four seismic activity regions. In our current analysis, we observe that nearly the same regions have emerged. This indicates a convergence between our manually conducted analysis and the results obtained using the elbow method.

\begin{figure}[h]
    \centering 
    \includegraphics[width=0.5\textwidth]{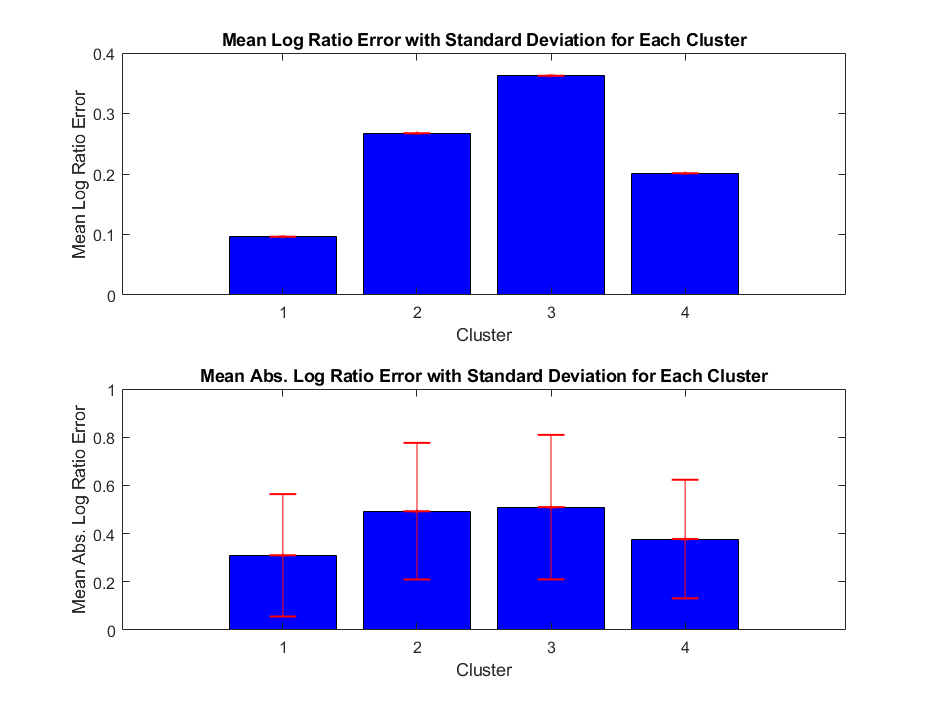} 
    \caption{Regional average $V_{s30}$ prediction errors} 
    \label{fig:my_label5} 
\end{figure}

In the following, the performance of the $V_{s30}$ prediction model in four different clusters was evaluated by calculating the average absolute log ratio error, which includes both the average log ratio error and standard deviations, as provided in Figure \ref{fig:my_label5}.

Positive values of the mean logarithmic ratio error (Figure \ref{fig:my_label5} upper graph) indicate that the predicted values are systematically higher than the measured values. The bottom graph presents the mean absolute logarithmic rate error including standard deviations for each cluster. This chart provides a clearer perspective on the consistency of prediction errors within each cluster. Cluster 1 exhibits significant variability in prediction errors, with a mean absolute log odds error of approximately 0.35 and a high standard deviation. Similarly, Cluster 3 also shows large variability, with a mean absolute log ratio error of approximately 0.45. In contrast, Cluster 2 and Cluster 4 provide more reliable and consistent estimates, with mean absolute log ratio errors and lower standard deviations of approximately 0.25 and 0.20, respectively. 

%Approximately 24\% of the earthquake records in the data set were taken from stations where $V_{s30}$ values were not measured. 
% The majority of the recordings were made on sites belonging to C and D soil classes, with approximately 41\% belonging to C and 22\% D class. With this information, regional average $V_{s30}$ prediction errors, focusing only on classes C and D, were calculated and shown in Figure 22.

% \begin{figure}[t]
%     \centering 
%     \includegraphics[width=0.5\textwidth]{regionalavg.png} 
%     \caption{Regional average $V_{s30}$ prediction errors of stations belonging to C and D soil classes} 
%     \label{fig:my_label6} 
% \end{figure} 

%Figure \ref{fig:my_label6} shows the average absolute logarithmic ratio error in blue, Class C logarithmic ratio errors in yellow, and Class D logarithmic ratio errors in black. It is observed that the $V_{s30}$ predictions can be regarded as relatively successful for sites that fall into site class C category. Stations of site class C dominate the data set. Conversely, the predictions are less accurate for stations in the D soil class, where data availability is relatively limited. \\

\section{Future Directions}

In this study, we aimed to extract both local and time-dependent features from time-series data by using a hybrid architecture that combines CNNs and LSTMs. This study builds on our previous studies where we employed supervised methods to improve $V_{s30}$ predictions. By integrating CNNs, which are effective for capturing local patterns, with LSTMs, which model temporal dependencies, we successfully enhanced the predictive performance of our models.

As a future direction, an unsupervised approach for time-series signals, similar to those used in NLP, appears highly promising. Text data, like time-series data, exhibits sequential characteristics with both short- and long-term dependencies. Recent advancements in NLP, such as transformer-based models and other unsupervised techniques, have shown great success in extracting meaningful patterns from text data. Extending these methods to time-series signals, including seismic data, could provide significant improvements.

The ongoing development of methods for time-series analysis in the literature suggests that seismic signals could also benefit from these approaches. Adapting unsupervised learning techniques to earthquake signal analysis has the potential to open new avenues for understanding and predicting seismic phenomena more effectively.

\section{Acknowledgments}
This study was supported by the Scientific and Technological Research Council of Turkey (TUBITAK) under the Grant Number 121M732. The authors thank TUBITAK for their support.

\bibliographystyle{plainnat}
\bibliography{refs}

\end{document}